\documentclass[twoside,leqno,twocolumn]{article}  
\usepackage{ltexpprt} 
\usepackage{graphicx}
\usepackage{amssymb}
\usepackage{textcomp}
\usepackage[backend=bibtex]{biblatex}

\begin{document}

\title{\Large Tracking Motion and Proxemics using Thermal-sensor Array 
}
\author{Chandrayee Basu\qquad Anthony Rowe\\Carnegie Mellon University}

\date{2014}

\maketitle

\begin{abstract} \small\baselineskip=9pt 

Indoor tracking has all-pervasive applications beyond mere surveillance, for example in education, health monitoring, marketing, energy management and so on. Image and video based tracking systems are intrusive. Thermal array sensors on the other hand can provide coarse-grained tracking while preserving privacy of the subjects. The goal of the project is to facilitate motion detection and group proxemics modeling using an 8 x 8 infrared sensor array.  Each of the 8 x 8 pixels is a temperature reading in Fahrenheit. We refer to each 8 x 8 matrix as a scene. We collected approximately 902 scenes with different configurations of human groups and different walking directions. We infer direction of motion of a subject across a set of scenes as left-to-right, right-to-left, up-to-down and down-to-up using cross-correlation analysis. We used features from connected component analysis of each background subtracted scene and performed Support Vector Machine classification to estimate number of instances of human subjects in the scene.  

\end{abstract}

\section{Introduction}

Indoor tracking has pervasive applications beyond mere surveillance, for example in education, health monitoring, marketing, energy management and so on. Recent research has shown that knowledge of occupancy alone can save 42\% annual air conditioning energy savings in commercial buildings \cite{Erickson14}. Erickson \textit{et al.} also showed that binary Passive Infrared (PIR) sensing is inadequate for building controls and energy savings. Knowing the actual room usage and occupancy patterns can significantly influence how efficiently a control system can balance energy savings considerations and thermal comfort \cite{scott2011},\cite{koehler2013},\cite{balaji2013}. \\

Rooms are scheduled for meetings and classes assuming attendance of all registered occupants or participants, resulting in inefficient space scheduling and usage. Setting the wrong occupancy-based temperature has a negative impact on either energy savings or thermal comfort. Thus knowing how many people occupy a room allows for a reduction of ambient temperature in cases of high occupancy or an increase in cases of low occupancy. Despite the obvious benefits of having a reliable real-time occupancy count, very few building systems have a measure of occupancy beyond simple binary presence-absence. Predicting occupant locations and when they will transition between these locations is challenging.Video based tracking can provide fine-grained information for such predictions. One simple application of such tracking would be to mount some imaging device at the doorway and count how many people enter and leave the space at a given point of time.\\

Beyond efficient building service usage and energy management, tracking a group of people in a social setting has other applications. Group behavior like touch and proxemics vary between countries. The touch sensitivity of a child with ASD is different from that of a normal person. Coarse information about activities within a group speaks of group dynamics, which may be valuable piece of information in applications that encourage group formation and sociopetal behavior. \\

Several methods have been proposed for detecting, counting, tracking, and identifying people inside buildings using a combination of different sensing modalities and machine learning approaches \cite{Trumler2005}. Image and video based tracking systems are intrusive. Thermal array sensors on the other hand can provide coarse-grained tracking while preserving privacy of the subjects. The goal of the project is to facilitate motion detection and group proxemics modeling using an 8 x 8 infrared sensor array. 

\section{Related Work}

Estimate number of instances: Erickson used thermal sensor array to estimate people count in the field of view of a thermal sensor array \cite{Erickson09a, Erickson11a, erickson2013}. The researchers extracted three features viz. number of connected components, size of the biggest connected component and number of active pixels from Connected Component Analysis of 8 x 8 pixels temperature data. This method produced plausible results except when the members of the group in the scene are not too close to be indistinguishable as separate temperature blobs. We found that local excitation peaks within the connected components carry more information about the actual number of people in the scene, in addition to the above features used by Erickson \textit{et al.} Moreover with labeled data it should be possible to perform supervised learning rather than clustering. In a later paper \cite{bertrand2013} the authors used Artificial Neural Network and K-Nearest Neighbor Classification and regression and applied filtering on the estimations to determine number of people in the scene. Support Vector Machine (SVM) is used to classify facial images by pose as an alternative to Neural Network (NN). In fact SVM can be used to replace several applications of NN. In this work using SVM in place of ANN or linear regression reduces the amount of noise in the estimated number of instances, therefore obviating the filtering.

Motion tracking: Several simple deterministic to probabilistic algorithms are used for motion tracking. Probabilistic methods like Kalman Filter, Particle Filter or general Hidden Markov model are more viable and relevant approaches when the sensor array consists of sensors that are sufficiently apart to allow for uncertainties in the process of movement. 

Normalized and Generalized  cross correlation is a simpler method that has been used for motion tracking from images. Normalized cross correlation (NCC) is the most robust correlation measure for determining similarity between points in two or more images providing an accurate foundation for motion tracking \cite {ncc}. Hii \textit{et al.} used NCC for detecting motion of breast during Digital Image-based Elasto-tomography for Breast Cancer. Cross-correlation is extensively used in such sequential pattern recognition in neuroscience. The functional connectivity between neurons can be detected using cross-correlograms \cite{neuralact}. We used cross-correlation between the time series of pair of pixels and thresholded delay between neighboring pixels to estimate the direction of motion.  

\section{Problem Definition}

\begin{figure} \label{fig:1}
\centering
\includegraphics[width = 8cm]{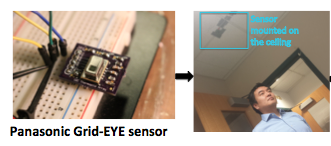}
\caption{Data acquisition set up showing Grid-EYE sensor array}
\end{figure}

System overview: The sensor used in this study is an 8 x 8 2D array of infrared sensors marketed as GridEYE by Panasonic \cite {grideye}. The infrared sensors are thermopile-type which detects quantity of infrared ray. These high precision sensors, based on advanced MEMS technology, have temperature detection, achieved on a two dimensional area, digital output. The horizontal and vertical angles of view are 60 deg each. The array has a floor coverage of 2.5 m x 2.5 m when installed 3 m above the ground. The sensor-array is mounted to a FireFly node \cite{firefly}. Figure 1 shows the data collection system configuration.

Data: The 64 temperature readings are quantized and range from range from –4\textdegree F to +212\textdegree F at low gain. The sampling rate is 10 frames/second. Human walking speed varies from 2.5 meters/second to 3 meters/second. Hence with 10 frames/second it is possible to capture motion from excitation of consecutive pixels with the array. However, faster motion like running or even hurried walking may not be detectable easily. 

The static background scene captured during the experiments was found to have a narrow range of temperature the standard deviation being 5\textdegree F.

Our dataset has 902 scenes which consists of the following:

\begin{itemize}
\item 4 different configurations of a single person scene with a tall and short person, each capture consisting of 656 scenes. 
\item 4 different configurations of two persons scene with different levels of closeness and different heights of the people in the group, each capture consisting of 656 scenes.
\item 2 different configurations of three persons and four persons scene with people of different heights, each capture consisting of 656 scenes.
\end{itemize}

We took the above measurements at the doorway of a 15 people lab with variable traffic throughout the course of the day. The measurements have all been controlled thus far.

\section{Methods} \label{Methods}

\subsection{Feature extraction} \label{algos}

Background subtraction: We collected 164 samples of background scene without human being. We extracted the foreground scene from frames with people by subtracting per pixel long term average temperature as shown in equation \ref{eq:1}. 

\begin{equation} \label{eq:1}
BS_{i,j} = P_{i,j} - \frac{1}{N}\sum_{k = 1}^N B_{i,j}^k 
\end{equation}

$BS_{i,j}$ is the background subtracted $(i,j)^{th}$ pixel value. $P_{i,j}$ is the value of the same pixel when there is human in the scene. The average temperature per pixel taken over 680 background frames was found to be 96 \textdegree F to 98\textdegree F. This is much lower in comparison to human skin temperature, the maximum we recorded being 106\textdegree F to 113 \textdegree F near the head for people of varying heights.

\begin{figure} \label{fig:2}
\centering
\includegraphics[width = 5cm]{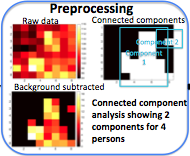}
\caption{Raw data with 4 people in the scene, background subtracted matrix and connected component matrix (counter-clockwise)}
\end{figure}

Connected component analysis: Connected component analysis or labeling is also called blob extraction and commonly used in computer vision problems. It is an algorithmic application of Graph Theory in which subset of the connected components are uniquely labeled based on a heuristic. In our case the algorithm starts from left top pixel of the image and labels any pixel that does not have a formerly encountered connected component before, above and diagonally above it. The above is called the $Two-pass$ method for blob extraction. From our dataset we found that due to the coarse level of the data, it is possible to extract number of people in the scene just from the number of connected components in a background subtracted image if the members in the scene are far apart and are standing casually without moving hands. See Figure 2 for an example of a four people scene after background subtraction and connected component analysis. 

\begin{figure} \label{fig:3}
\centering
\includegraphics[width = 8cm]{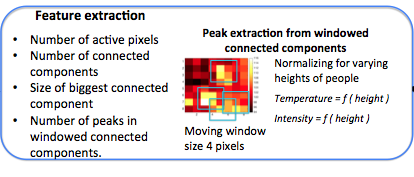}
\caption{Feature selection with local peak detection using 4 pixels sliding window}
\end{figure}

Local peak count: Human skin temperature is usually the maximum at the head and chest. Therefore any individual standing under the infrared array with appear to have a local high temperature pixel and surrounding lower temperature valleys. When several people are close together in the scene it is not possible to separate the number of people from the blobs or even size of the blob. We found that detecting local peaks within the connected components and the number of such peaks per scene is an useful feature for number of instances estimation. The local peak count estimation is illustrated in Figure 3.

The number of local peaks detected depends of the height of the person.We used a sliding window of size 4 pixels. Two neighboring high temperature pixels within one window indicates a tall person. This is because the amount of the infrared ray decays by square of distance with increasing distance from the sensor. The algorithm below assumes that temperature or intensity recorded at the sensor array is proportional to the height of the person. The peak detection algorithm is as follows:\\

//Start from the top left pixel in the frame.//\\ 

$count$ = 0

At each window: 
If $BS_{i,j}$ $>$ $\frac{global max}{1.2}$\\
$count$ = $count$ + $1$ \\

//Correction for tall people//\\

$totalcount = totalcount*correctionfactor$ \\

\subsection{Number of instances estimation}

The feature vector $X$ is 4 dimensional consisting of number of active pixels after background subtraction, number of connected components, size of the biggest connected component and number of local peaks in each scene. We performed K-means clustering using the first three features following Erickson \textit{et al.} \cite{erickson2013} and compared the with the case when all 4 features are used. However, since we are using labeled data a supervised learning approach would be more relevant for the available dataset. In order to estimate the number of people in the scene we used Support Vector Machine classification of the feature space. In instance estimation work, the goal is to classify each scene as having 1, 2, 3 or 4 instances. Since the number of people that can gather in 2.5 m x 2.5 m space is utmost 5, we restricted the number of classes to four. The class labels are $y \in {1,2,3,4}$ 

Support Vector Machine (SVM) classifies instances by minimizing structural risk. Linear SVM is a large margin classifier, the objective function is to maximize the width of the margin without hitting the data points (Also called support vectors). The standard SVM problem is written as \cite{svm}:

\begin{equation}
argmin \frac{1}{2} \textbf{w}^T\textbf{w} + C \sum_{i = 1}^l \xi_i
\end{equation}

subject to $y_i(\textbf{w}^T\phi(\textbf{x}_i) + b) \geq 1 - \xi_i, \xi_i \geq 0, i = 1,...,l$\\

$\phi(\textbf{x}_i)$ is the mapping of $\textbf{x}_i$ to a higher dimensions for non-linear classification problems. $\textbf{w}$ is a vector in higher dimensional space. The problem is solved using the Lagrangian Dual formulation. $C$ is the penalty parameter for error tolerance. We used Radial Basis Function kernel for the classification problem. The parameters of SVM model are $C$, kernel type and kernel width $\gamma$. We divided the dataset into training feature vector $X \in \mathbb{R}^{600\times 4}$and testing feature vector $Xtest \in  \mathbb{R}^{220\times4}$ and performed 10-fold cross validation on the training data for parameter selection. The training and testing data consisted of varying configurations of 1,2, 3 and 4 persons in the scenes. 

\subsection{Motion tracking}

Each 8 x 8 pixels scene captures motion at a coarse level. The field of view being only 2.5 m x 2.5 m the estimation of direction of motion of one person can be captured fairly easily by analyzing the sequential pattern of exciting of the cells in the matrix. For several scenes with motion we extracted the time series from each pixel and computed a cross-correlation matrix for each cell with 63 other cells. Figure 4 shows the time series of 5 consecutive cells. 

\begin{figure} \label{fig:4}
\centering
\includegraphics[width = 8cm]{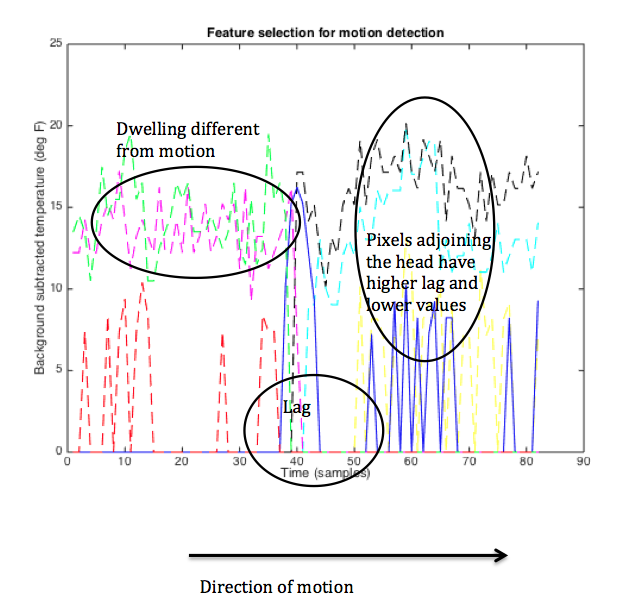}
\caption{Time series data of 5 consecutive pixels along the $4^{th}$ column of the grid}
\end{figure}

\section{Results}

\begin{figure} \label{fig:5}
\centering
\includegraphics[width = 8cm]{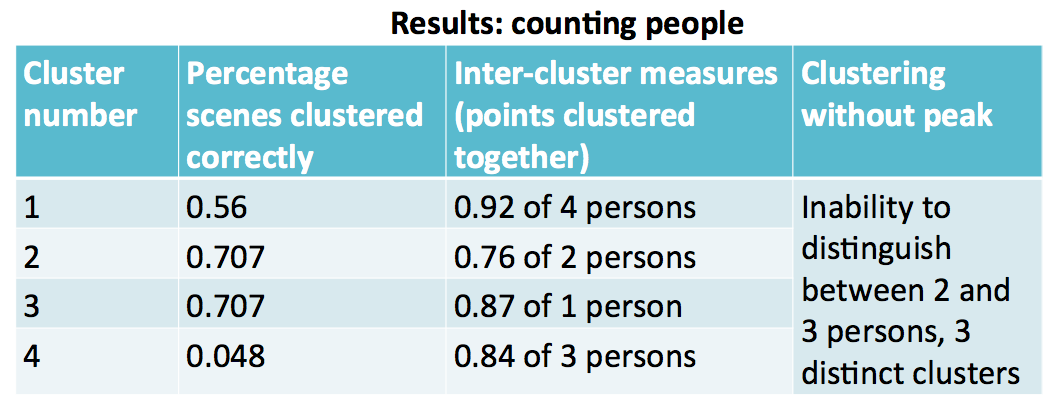}
\caption{Result of K-means clustering}
\end{figure}

\begin{figure} \label{fig:6}
\centering
\includegraphics[width = 8cm]{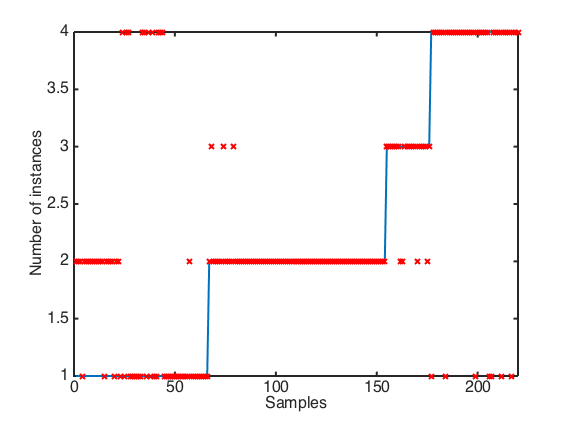}
\caption{Result of Support Vector Machine Classification on mixed labeled data, the red cross markers show the estimated labels}
\end{figure}

\begin{figure} \label{fig:7}
\centering
\includegraphics[width = 8cm]{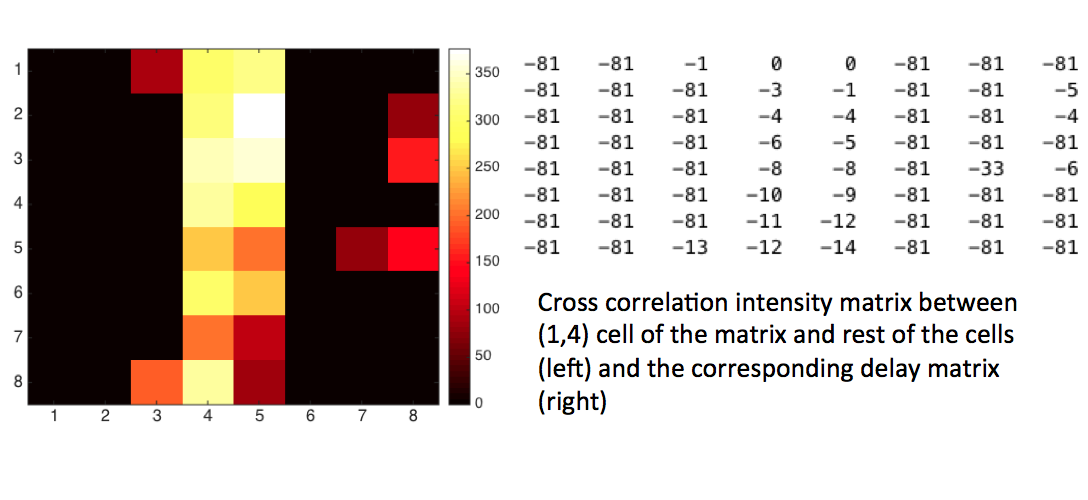}
\caption{Cross-correlation matrix between cell(1,4) and all cells (left) and corresponding delay matrix (right), direction of motion down-to-up}
\end{figure}

\subsection{Number of instances estimation}

The results of K-means clustering using 3D and 4D feature vectors are shown in Figure 5. The cluster number indicates the estimated number of occupants. From the results we can only infer that 92 \% of all 4 people scenes are grouped together. Similarly 76 \% of 2 people scene and 87 \% of all the 1 people scenes are grouped in two different clusters. 84 \% of 3 persons scenes are grouped together during training. These results are from 4 D feature vector. However from this result it is difficult to say which cluster corresponds to which count. Using 3 D feature vector as in Erickson \textit{et al.} \cite{erickson2013} two people and three people scenes were clustered together.

Next we present the results of SVM classification of the scenes by the number of instances (persons in this case). The best set of parameters using 10-fold cross validation were found to be: $C$ = 21, $\gamma$ = 0.0078. The resultant parameters were used to train on entire $X$ data. The training correct classification accuracy 89.8 \%. The testing accuracy was found to be 80 \% (see Figure 6 for results of testing.The red cross markers show the estimated labels against the blue ground truth labels. 

\subsection{Motion tracking}

A sample cross-correlation matrix and delay matrix are shown in Figure 7. The figure shows cross correlation intensity between the cell(1,4) and rest of the cells. The direction of motion is from down-to-up as indicated the delay between cell (1,4) and cell(8,4) and cell(8,5). We set a global threshold on the delay and the cross-correlation values between consecutive cells. We estimated the lag between adjoining cells for three walking scene sequences. The lag varied between 1 and 2 samples. At a sampling speed of 10 samples/second, this corresponds to 14 samples. The area of coverage being 2.5 m x 2.5 m, the estimated speed on walking within the field of view of the sensor matched the ground truth walking speed. From the activation pattern of the cells we can present the direction of motion.

\section{Discussion}
The features extracted from connected component analysis and the peak count together are more discriminative of the number of instances than the connected component features alone. This is evident from the results of K-means clustering. Using Support Vector Machine classification on the dataset, it is possible to identify most of the number of instances correctly unlike in K-means. We found the accuracy of estimation on test data to be 80 \%. For motion tracking cross-correlation and delay between boundary cells can be good indicators of direction of motion. This problem is being made easier by background subtraction. Note that the delay for most of the inactive cells is -81. Activation of adjoining cells which are not necessarily seeing the person walking can make the task difficult. However, a delay threshold of 2 samples between adjoining cells is sufficient for estimating direction of motion from the current dataset. We have not applied the method to more complex scenes, like one person walking into the scene while there are several people in the scene.

\section{Conclusion}
In this paper we developed low cost methods to estimate number of people and direction of motion within the field of view of a 64 pixels infrared sensor array. The output of each sensor is a temperature reading in Fahrenheit. We collected approximately 902 scenes with different configurations of human groups and different walking directions. Using Support Vector Machine classification on connected component based features and local peak counts we estimated the number of instances (for 1 to 4 instances) with 80 \% accuracy. We inferred direction of motion of a subject across a set of scenes as left-to-right, right-to-left, up-to-down and down-to-up using cross-correlation analysis.

\section{Future work}

Current data acquisition set up only allows for four directions of motion. In future probabilistic approach like Markov Random Fields could be used to estimate direction of motion across more complex scenes. The results can be applied to model and infer daily routines and room occupancy.

\section*{Acknowledgements}

We acknowledge Professor Anthony Rowe of Electrical and Computer Engineering, Carnegie Mellon University for providing the sensor and several members of Ubicomp Lab for their assistance in data acquisition.


\printbibliography

\end{document}